# RoboTacDex: A Dexterous Visual-Tactile-Action Dataset for Humanoid Manipulation

Xinyi Wang[1], Donghan Li[1], Zi'Ang Chen[1], Chong Yu[1], Chen Xin[2], Peng Ye[3,4], Yingkai Sun[1], Tao Chen[1]

***Abstract*—In the field of robot learning, large-scale and diverse demonstration trajectories provide the fundamental basis for enhancing robotic manipulation ability. We introduce RoboTacDex, a large, multi-modal, and diverse dataset of dexterous manipulation behaviors performed with a humanoid robot. Built on the publicly accessible humanoid robot Unitree G1, RoboTacDex consists of 6k trajectories covering 19 tasks, 23 skills, and interactions with 22 objects. RoboTacDex provides comprehensive records including multi-view RGB and depth information, tactile feedback, and detailed semantic annotations. Furthermore, the dataset features a variety of relatively challenging tasks that can only be completed by dual arms and dexterous hands, aiming to mimic human-like operational logic and simulate real-world manipulation complexity. To ensure data collection quality, we develop an improved multi-camera synchronization system to enable millisecond data synchronization and recording of modalities. In our experiments, we evaluate three representative imitation learning models on our dataset, analyzing their performance as well as their respective strengths and limitations across different task categories. Successful trial results and a moderate level of generalization capabilities across a suite of tasks indicate the effectiveness and diversity of the collected dataset. Our dataset will be open-sourced soon.**

## I. INTRODUCTION

In recent years, imitation learning and related methodology leveraged by large-scale datasets have achieved rapid progress in the field of robotic manipulation. Data collection systems for fixed-base robots have become relatively mature and have yielded numerous large-scale datasets [1]–[3]. These efforts have effectively driven the progress of various manipulation policies [4]–[6]. However, high-quality manipulation datasets specifically for humanoid robots remain extremely scarce. Compared to fixed-base manipulators, data collection for humanoid robots faces challenges. On one hand, humanoids have extremely high degrees of freedom (DOF), requiring collection systems that can precisely record complex, high-dimensional bimanual coordination tasks. On the other hand, there are high expectations for humanoids to perform complex tasks in unstructured, open-world environments, which sets a higher bar for data diversity, system completeness, and the objectivity of evaluation frameworks.

Currently, the surge in robotic manipulation research such as diffusion model and Vision-Language-Action (VLA) models has demonstrated immense potential for developing general-purpose robotic manipulation models. However, pure visual input from a single viewpoint is often insufficient for handling complex physical interactions and achieving real generalization. Multi-view observations and depth information provide a complete spatial geometric representation [7], while tactile information serves as a core feedback mechanism for achieving dexterous manipulation and perceiving the physical properties of objects [8], [9]. Consequently, constructing a humanoid manipulation dataset that integrates multi-modal information, including vision, depth, and tactile sensing, is a critical cornerstone toward achieving universal embodied intelligence.

To bridge this gap, this study presents RoboTacDex, a large-scale, multi-modal dataset specifically targeting dexterous upper-body humanoid manipulation. This dataset comprises over 6,000 demonstration trajectories, totaling approximately 25 hours. Distinct from earlier research confined to simple tabletop tasks or single modalities, this dataset provides comprehensive records including tactile feedback, multi-view depth maps, and RGB images. The tasks are designed to simulate authentic human-like operational logic, demonstrating a wide range of skills and task complexity. To ensure the multimodal temporal consistency of the collected trajectories, we design a hardware-software co-synchronization method. Furthermore, we re-engineer the Unitree official teleoperation script to enable the recording of depth and tactile sensing data.

Furthermore, this work not only provides data resources but also establishes a verification benchmark. Through extensive experimental analysis, we verify the significant role of this dataset in enhancing robotic policy performance. We also evaluate the performance of representative imitation learning policies when handling high-dimensional humanoid action spaces, with experimental results proving the importance of high-quality, multi-modal data for breaking through current bottlenecks in humanoid manipulation tasks.

The main contributions of this paper are summarized as follows:

- We construct a multi-view, multimodal humanoid upper-body manipulation dataset containing over 6k trajectories and covering diverse human-like tasks.
- We develop an improved hardware integration scheme that resolves synchronization challenges across heterogeneous sensors.
- We conduct evaluations of representative manipulation learning methods on the RoboTacDex dataset, providing a standardized performance benchmark and technical reference for future humanoid manipulation research.

[1]Fudan University; [2]ByteDance Intelligent Creation; [3]Shanghai AI Laboratory; [4]The Chinese University of Hong Kong.

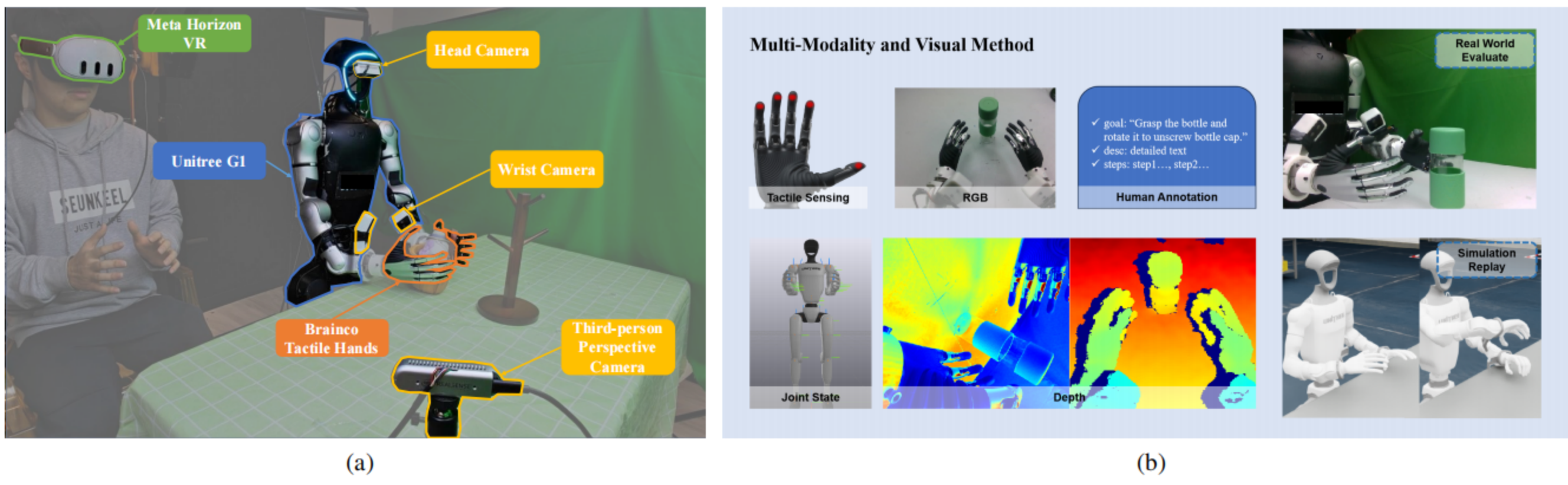

(a) (b)

Fig. 1. RoboTacDex Overview. (a) The framework of teleoperation system. (b) RoboTacDex aggregates high-quality multi-modal sensory data and provides two method to visualize. We can replay the record trajectory in real world robot and IsaacSim simulation environment to ensure correct collection.

## II. Related Work

### A. Robot Policy Learning

Traditional robot manipulation systems predominantly rely on model-based planning and control framework. This indicates that systems often struggle to generalize to unseen scenarios. In contrast, imitation learning policies enable robots to acquire stable manipulation skills by imitating expert behavior [4], [10], [11]. Recent advances in diffusion-based models [12] have achieved remarkable success in image generation and beyond. Building on these developments, diffusion policy [5] has been introduced to robotic manipulation tasks, transforming action generation into a process of progressively denoising. Many small-scale models [13]–[16] have been proposed and have achieved promising results on specific tasks. Their generalization capability heavily relies on sufficient and various data. Moreover, occlusions during data collection often compromise task success. Fusing multi-view visual information provides richer spatial cues [7], [17] and mitigates occlusion issues. However, multi-view RGB observations still lack explicit geometric structure. Incorporating explicit depth data or 3D representations [6], [18]–[21] provides robots with a more comprehensive geometric structure feature. The incorporation of depth information reduces the need for thousands of demonstrations to implicitly model 3D geometry feature in 2D model. Previous research has also demonstrated that multi-view or 3D representation-based policies [22]–[25] significantly outperform models relying solely on single-view or 2D images in task success rates and generalization to scene distribution shifts. And tactile sensing is equally crucial in robotic manipulation, compensating for visual occlusions and providing finer perception or feedback from manipulated objects in complex tasks [9]. Several studies [8], [26]–[28] have attempted to integrate tactile sensing with imitation learning.

More recently, VLA models [25], [29]–[31] building upon multimodal large language models (MLLMs), exploit their strong representation capacity and commonsense reasoning about the physical world. These models are pretrained on vast-scale web data and implicitly acquire rich semantic and spatial priors for manipulation. Many existing VLA models [32], [33] demonstrate strong versatility and generalization across diverse scenarios, objects, and even robot types.

### B. Robotic Manipulation Datasets

Robot manipulation datasets play a crucial role in facilitating the learning of data-driven robot policies. For years, robotics researchers have invested significant time and effort in constructing high-quality supervised datasets. Given the advantages of simulation environments in scalability, controllability, and data acquisition costs, researchers have primarily focused on building operational scenarios within simulations to create datasets [4], [10], [34]–[36]. However, due to persistent, non-negligible differences between simulation and the real physical world in dynamics and contact modeling, reliance solely on simulated data often suffers from the sim-to-real gap. Datasets like RoboTurk [1], BC-Z [37], RT-1 [38], BridgeData [2], and RH20T [39] collect data in real environments through human teleoperation or scripted methods. Building on the success of the VLA paradigm in manipulation, the robotics community has proposed large and diverse datasets aimed at improving generalization in varied environmental conditions. Researchers have integrated cross-robot large-scale multi-task datasets [40], [41].

Recently, humanoid robots have garnered significant attention, spurring a surge in the development of humanoid-specific datasets. Due to the relative immaturity of teleoperation for humanoids compared to robotic arms, data collection suffers from suboptimal control precision and efficiency. The simultaneous necessity of stabilizing the lower body further complicates the acquisition of humanoid manipulation datasets. The data collected by Nvidia Gr00t [33] contains 240k table manipulation trajectories, but it records only one RGB camera. RoboMIND [3] is a multi-entity dataset containing approximately 15k humanoid trajectories, but it relies on a single camera for RGB and depth recording, with no tactile data captured. To the best of current knowledge, Agibot World [42] is the largest humanoid manipulation dataset, employing multi-angle standard RGB and fisheye cameras for supplementary views. However, dexterous hand data constitutes only a small fraction, and no tactile information. Humanoid Everyday [43] records rich multimodal information including visual, tactile, and textual data, yet it also lacks multi-view coverage. In contrast, our dataset distinguishes itself by compiling comprehensive multi-modal data, encompassing RGB, depth, tactile inputs and natural language annotations (see Table I). It records the whole manipulation process from 4 perspectives, offering a valuable resource for rich physical interactions tasks and multi-view alignment research in robot manipulation.

TABLE I
COMPARISON OF ROBOTIC MANIPULATION DATASETS

| Dataset | # Traj. | # Skills | Humanoid Robot | Dexterous Hand | Human-Robot Interaction | Tactile Sensing | Multi-view Camera |
|---|---|---|---|---|---|---|---|
| RH20T [39] | 13k | 33 | ✗ | ✗ | ✗ | ✓ | ✓ |
| BridgeData [2] | 7.2k | 4 | ✗ | ✗ | ✗ | ✗ | ✓ |
| DROID [40] | 76k | 86 | ✗ | ✗ | ✗ | ✗ | ✓ |
| Open X-Embodiment [41] | 1400k | 217 | ✗ | ✗ | ✗ | ✗ | ✓ |
| Fourier ActionNet [44] | 13k | 16 | ✓ | ✓ | ✗ | ✗ | ✗ |
| RoboMIND [3] | 107k | 38 | ✓ | ✓ | ✗ | ✗ | ✗ |
| Agibot World [42] | 1000k | 87 | ✓ | ✗† | ✓ | ✗ | ✓ |
| Humanoid Everyday [43] | 10.3k | 221 | ✓ | ✓ | ✓ | ✓* | ✗ |
| **RoboTacDex** | **6k** | **23** | ✓ | ✓ | ✓ | ✓ | ✓ |

† *Mostly grippers and only a few dexterous hands.* * *Only three-fingered (Dex-3) hand with tactile perception.*

## III. DATA COLLECTION PLATFORM SETUPS

### A. Hardware Design

In this section, we describe the setups of the entire data acquisition system. As shown in Fig. 1a, the operator wears a Meta Horizon VR to capture human upper-body movements and map them onto the robot, controlling the motion of each joint of the arms and fingers. We employ a Unitree G1 humanoid robot for data collection: dual arms with 14 DOF and dual Brainco dexterous hands with 12 DOF. The lower limbs and waist remain fixed to stabilize the robot in place, ensuring the head-mounted camera consistently captures desktop operations. Additionally, four RGB-D cameras capture multi-view observations: one provides a head-centric view, two are mounted on the wrists to focus on the dexterous hands, and one offers a third-person perspective overview of the operational scene. The dexterous hands are Brainco Revo2 Tactile version, capable of recording touch information such as normal forces and tangential forces on each finger.

### B. Data Collection Pipeline

We record trajectories on high-frequency 30Hz. Each frame includes low-dimensional state data for the arm and finger joints, RGB and depth information from four cameras with 640×480 resolution, tactile information from both hands, and action data of the arm and finger joints, as shown in Fig. 1b. Additionally, the collected data can be replayed in both simulated and real environments to ensure correct data recording.

*1) Camera Synchronization:* Robotic manipulation tasks exhibit significant temporal sequentiality and high dynamic characteristics. If observations from different perspectives are temporally misaligned, the same physical state may be erroneously modeled as inconsistent multimodal inputs. Therefore, timestamp alignment among multiple cameras is essential for ensuring geometric and semantic consistency across multi-view. In our data collection system, we employ a combination of hardware and software synchronization to minimize temporal mismatches among cameras capturing the same frame. The first-person and third-person perspective cameras are RealSense D435i, which support hardware synchronization. Meanwhile, software synchronization is applied between the head/third-person perspective cameras and the wrist-mounted cameras to eliminate temporal misalignment in the recorded video streams.

*2) Tactile Information Transmission:* The tactile sensing modality provides normal and tangential contact forces (including direction) at each fingertip, along with proprioceptive signals derived from self-capacitance proximity sensing— the latter serving as a measure of the distance between the finger and external objects. Tactile information and joint state data from the dexterous hand are published at a frequency of 100Hz in the DDS messages form. The local computer receives these messages and records them at 30Hz alongside other modalities.

## IV. DATASET ANALYSIS

Based on the data collection platform described in Section III, we construct a large, multi-modal, and diverse dataset named RoboTacDex. This dataset comprises 6k high-quality trajectories across 19 tasks, 23 skills, and 22 objects. In this section, we further analyze the dataset through a quantitative study of its diversity across task, object, and scene. And our evaluation across multiple dimensions shows that RoboTacDex constitutes a comprehensive dataset for manipulation policy learning and broader research.

### A. Diversity of Task and Skill Types

Different datasets contain diverse manipulation tasks, and a single task may involve several steps, different objects, and skills. To categorize these tasks, we define five common manipulation types, and each trajectory can belong to multiple manipulation types.

1) Basic Manipulations: Object interactions such as grasping, placing, or pushing without complex constraints.
2) Articulated Manipulations: Manipulations involving objects with articulated joints.

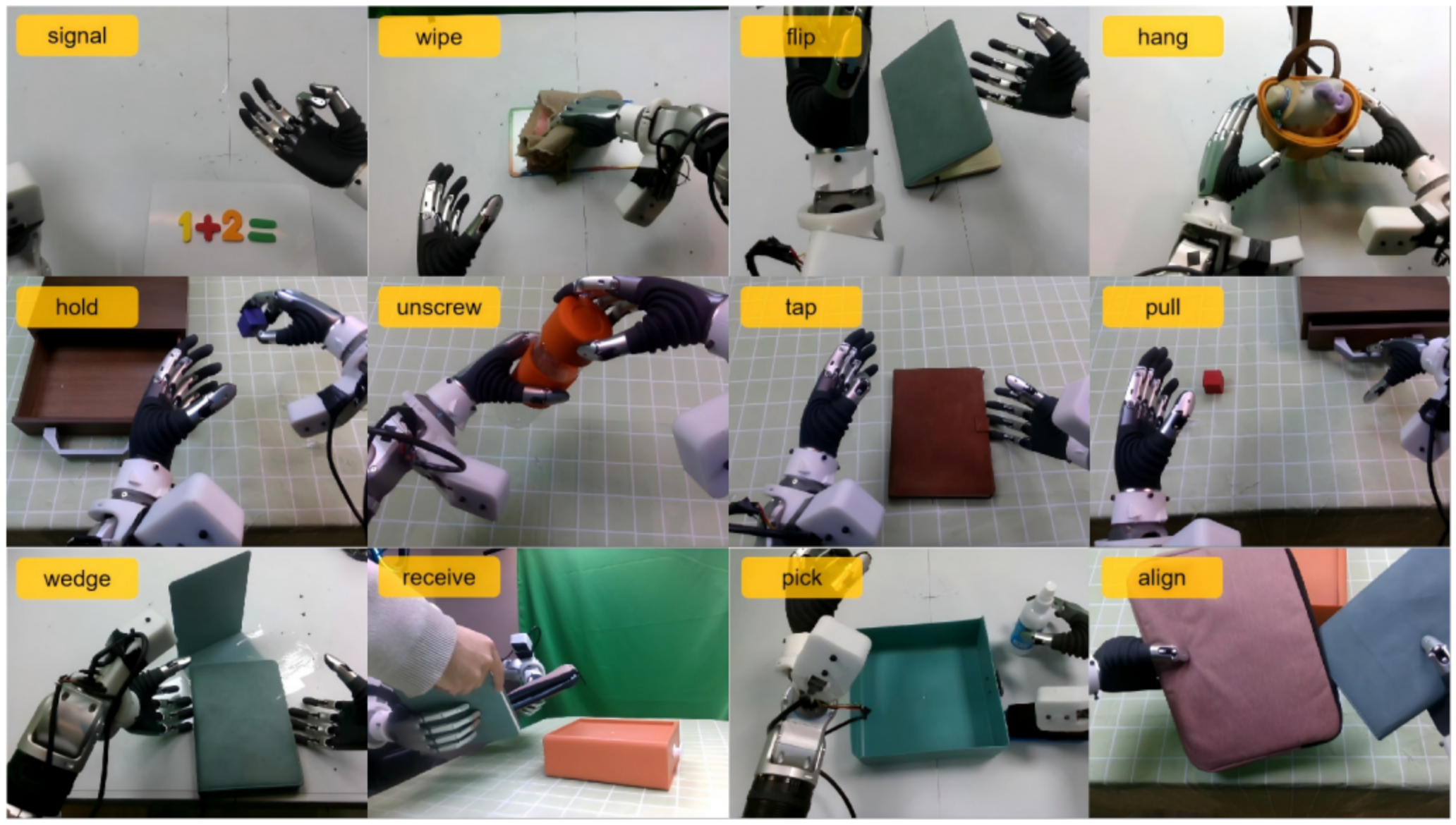

Fig. 2. Representative dexterous manipulation skills in RoboTacDex. The dataset comprises 6k high-quality trajectories across 19 tasks, 23 skills, and 22 objects, covering a wide spectrum of dexterous manipulation behaviors. And for many tasks, we collect data across 4 configurations: robot-to-table distances of 5cm and 15cm, combined with white and green grid background.

3) Dual-arm Collaborative Manipulations: Tasks requiring coordinated manipulation using two robotic arms.
4) Fine Manipulations: Precision-critical tasks requiring accurate contact control or fine-grained motion.
5) Humanoid Interactive Manipulations: Tasks related to the semantic understanding of the scene or requiring interaction with human.

Many existing datasets [2], [37] contain a large number of basic tasks, such as picking and placing objects in different scenarios. With these datasets, models have already mastered skills like pick-and-place with high accuracy, and for such tasks, a single-arm robotic arm is sufficient. We hope that RoboTacDex, as a dataset for dexterous hands on humanoid dual-arm robots, can demonstrate unique advantages in handling relatively complex scenarios. Therefore, when designing RoboTacDex, we defined multiple relatively challenging tasks that can only be completed by dual arms and dexterous hands (see Fig. 2), rather than being dominated by basic manipulations. The RoboTacDex dataset comprises 19 tasks and 23 skills, evenly distributed across diverse operation types. As illustrated in Fig. 3, numerous skills are not confined to basic manipulation but are widely utilized across various complex manipulation categories like dual-arm collaborative and fine manipulations. Conversely, each manipulation type integrates a rich set of skills, ranging from fundamental actions to specialized ones like unscrew and shielded grasp, ensuring that the dataset comprehensively covers a broad spectrum of human-like dexterous and multi-arm tasks. And the detailed atomic skills are shown in Fig. 4a.

### *B. Horizon, Object and Environmental Diversity*

Fig.4b illustrates the distribution of task durations across the dataset. The majority of trajectories are within the 10–40 s, indicating that most tasks require multiple coordinated actions rather than a single primitive or one-step execution. Meanwhile, a small portion of trajectories exhibit substantially longer durations. These tasks involve sustained manipulation and temporal causality, presenting a higher level of difficulty for data acquisition.

Similar to task types, we categorize objects involved in manipulations into five classes:

1) Articulated Objects: Objects with internal degrees of freedom that necessitate manipulation along specific trajectories or joint orientations.
2) Constrained Objects: Objects that can be grasped but require specific orientations or postures for manipulation.
3) Containers: Objects primarily used to support or hold other objects.
4) Functional Objects: Objects requiring functional interactions or coordination with other objects to complete tasks.

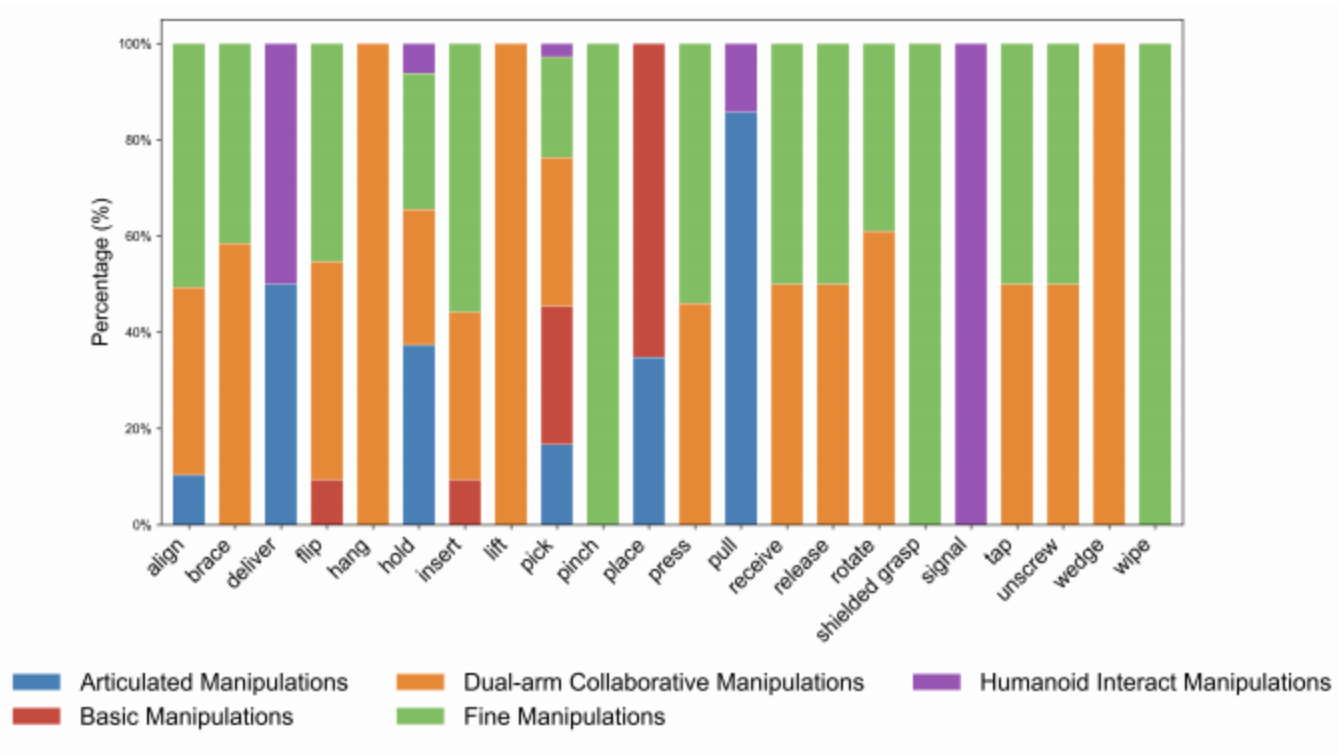

Fig. 3. Distribution of manipulation types and skills. Normalized proportion of skills utilized within each manipulation category in the RoboTacDex dataset, demonstrating extensive diversity and balanced distribution across various manipulation types.

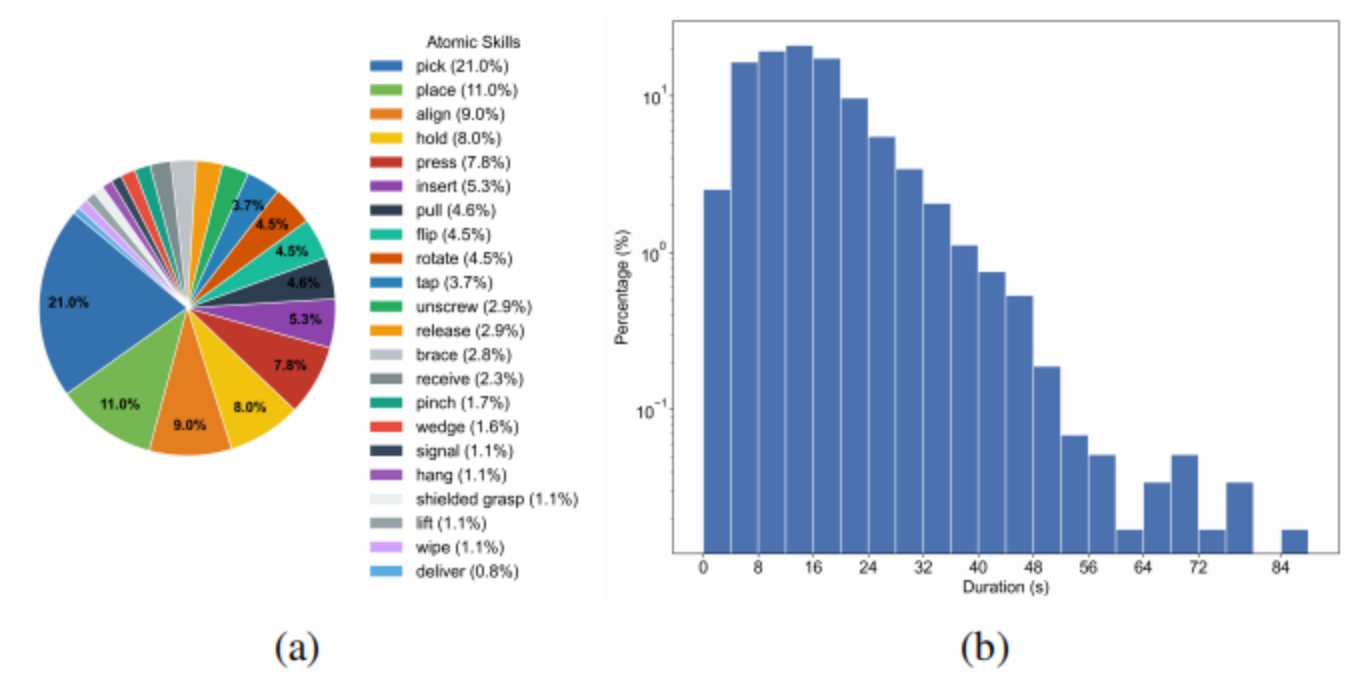

(a) (b)

Fig. 4. Dataset Features. (a) The various types of skills and their proportions in RoboTacDex (b) The horizon distribution in RoboTacDex.

5) Graspable Objects: Objects where primary manipulation goals can be achieved through a single grasp, without requiring specific manipulation directions or gestures.

As shown in Fig. 5, the dataset utilizes 22 distinct object types, ranging from kitchen food to office supplies, and encompassing both rigid and deformable objects with varied shapes, materials, and sizes. Notably, we ues the same object with different color, potentially providing valuable multiview video samples for data generation in robot manipulation.

Additionally, to investigate the impact of scene factors on task success rates, we control two environmental variables during dataset design: robot-to-table distance and table background. For most tasks, we collect data across 4 configurations: robot-to-table distances of 5cm and 15cm, combined with white and green grid backgrounds (see Fig. 2).

## V. Experiments

To validate the data quality of RoboTacDex and better understand the performance of existing imitation learning methods in dexterous manipulation tasks for humanoid robots, we evaluate three policies on RoboTacDex: Action Chunking with Transformers (ACT) [13], Diffusion Policy (DP) [5], and GR00T N1.5 [33]. We primarily conduct evaluations on the following four tasks, as shown in Fig. 6:

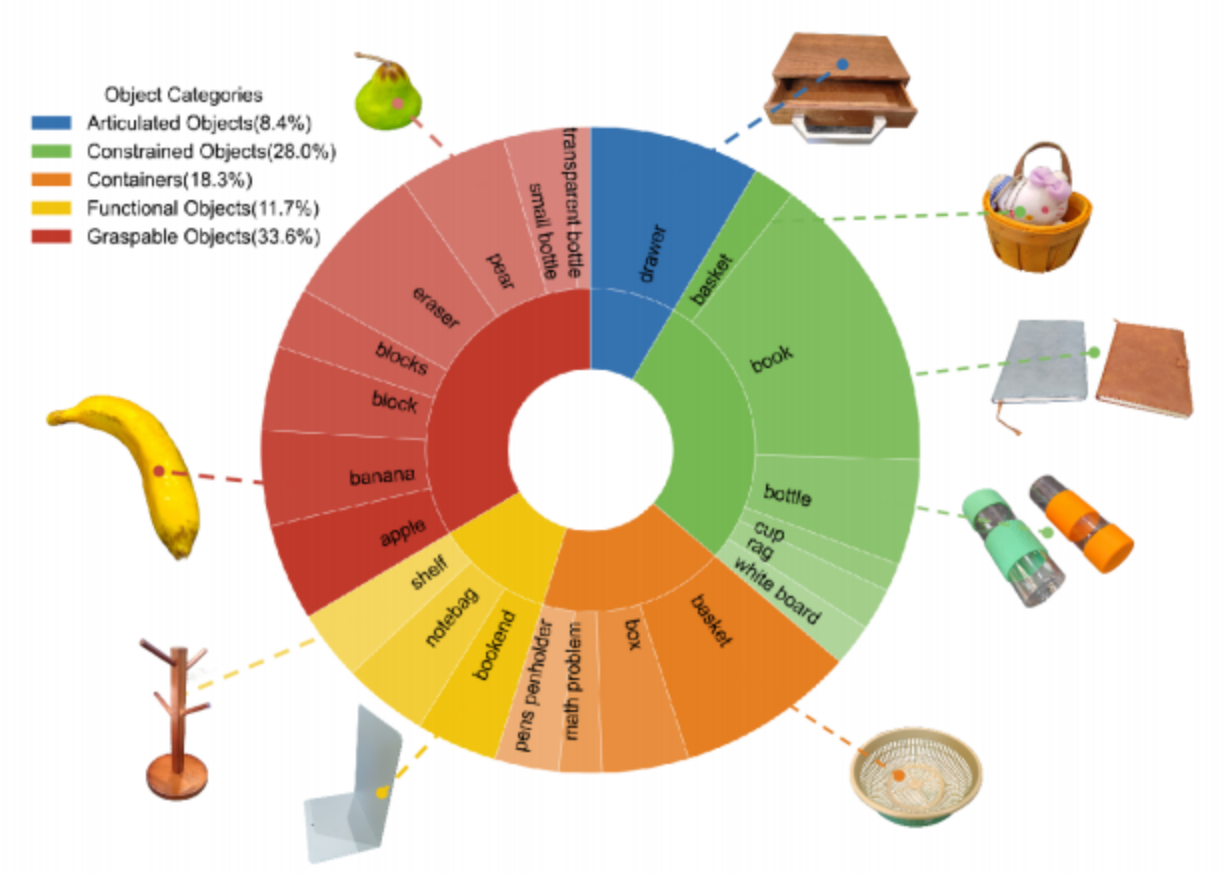

Fig. 5. The various objects and their types in RoboTacDex.

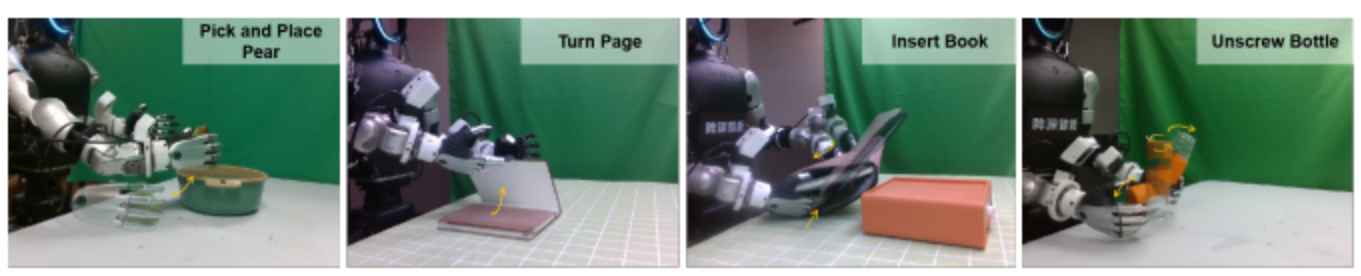

Fig. 6. Representative inference tasks to validate dataset. The yellow arrows show the robot's hand trajectories. For each task, we conduct 10 trials for different policies.

- `PickAndPlacePear`: Pick up the pear from the table and place it in the basket.
- `TurnPage`: Lift and turn the current page to reveal the adjacent page, ensuring the page is fully turned and lies flat on the opposite side.
- `InsertBook`: Hold the document bag with one hand and the book with the other, then insert the book securely into the bag.
- `UnscrewBottle`: Grasp the bottle firmly and rotate it counterclockwise to unscrew the cap from the bottle.

Based on our experimental experience, for each task, models trained on approximately 200 trajectories across the three architectures exhibit generally reasonable trajectory. This requires significantly more data compared to robotic arms, primarily due to humanoid robots and dexterous hands with up to 26 DOF and their complex action space. For each task, we train ACT, DP, and finetune VLA with the same dataset. We train these models until convergence, and during validation, we adjust each model's inference parameters to optimize performance. We use the baseline observation configuration, taking the head image and joint state as observations. The performance of these policies is summarized in Table II, with each policy evaluated over 10 trials per task.

TABLE II
Success Rates of Imitation Learning Methods on the RoboTacDex Dataset.

| Task | ACT | DP | GR00T N1.5 |
|---|---|---|---|
| `PickAndPlacePear` | 0/10 | 3/10 | 9/10 |
| `TurnPage` | 6/10 | 5/10 | 6/10 |
| `InsertBook` | 4/10 | 3/10 | 4/10 |
| `UnscrewBottle` | 3/10 | 2/10 | 6/10 |
| **Average** | **3/10** | **3/10** | **6/10** |

### *A. Performance of Imitation Learning Policies*

We find that VLA generally outperforms the other two models, primarily due to its base model pretrained on large-scale manipulation datasets. This enables strong generalization to object positions and manipulation distances beyond the training dataset, especially for tasks like `PickAndPlacePear` that are widely covered in existing datasets. In contrast, for uncommon tasks like `InsertBook` and `UnscrewBottle`, GR00T failed to learn critical semantic requirements and spatial constraints. During the inserting, it misaligned the book and document bag, leading to task failure. This further demonstrates that VLA requires substantial data to learn the critical manipulation keypoints for different tasks.

For the relatively smaller-scale models ACT and DP, the mechanism of their learning is mapping from observations to the action space. If the spatial positions of objects in the example data are scattered with large intervals, the success rate of interpolation points within the operational domain during inference is suboptimal. The target position of the output action resembles the most similar trajectory, but often deviates from the actual object's location.

Worth mentioning is that due to ACT's network architecture and temporary ensemble module, tasks involving multi-modality in the dataset (e.g., picking objects with left/right hand) may result in both hands executing actions but without reaching the target position during inference. The stochasticity and multi-modality of human demonstrations significantly increase the difficulty of model to learn observation to action mapping [13]. In contrast, by learning the gradient of the action score function, DP can better express multimodal action distributions and handle the diversity of example trajectories.

### B. Analysis of Multiview and Tactile Perception

To understand the impact of observation design on humanoid manipulation learning, we conduct ablation experiments by varying the visual viewpoints and sensory modalities while keeping the task, policy architecture, and training parameters.

We compare the success rates of three models under different observation configurations. In Fig. 7a, we visualize the performance on `PickAndPlacePear` and `InsertBook`. We observe that across both tasks, no model demonstrates significant performance improvement after adding wrist cameras or third-person perspective camera. For ACT and DP, the models lack mechanisms for multi-view image matching or additional supervision. Consequently, simply concatenating images from multiple viewpoints proves ineffective. This observation suggests that the benefits of multi-view observations are not automatically realized by existing manipulation learning frameworks, but rather depend on the model's ability to reason about geometric relationships across viewpoints [21], [45]. For GR00T, single-view success rates on simple tasks like `PickAndPlacePear` are already high. Additional views do not contribute to better performance.

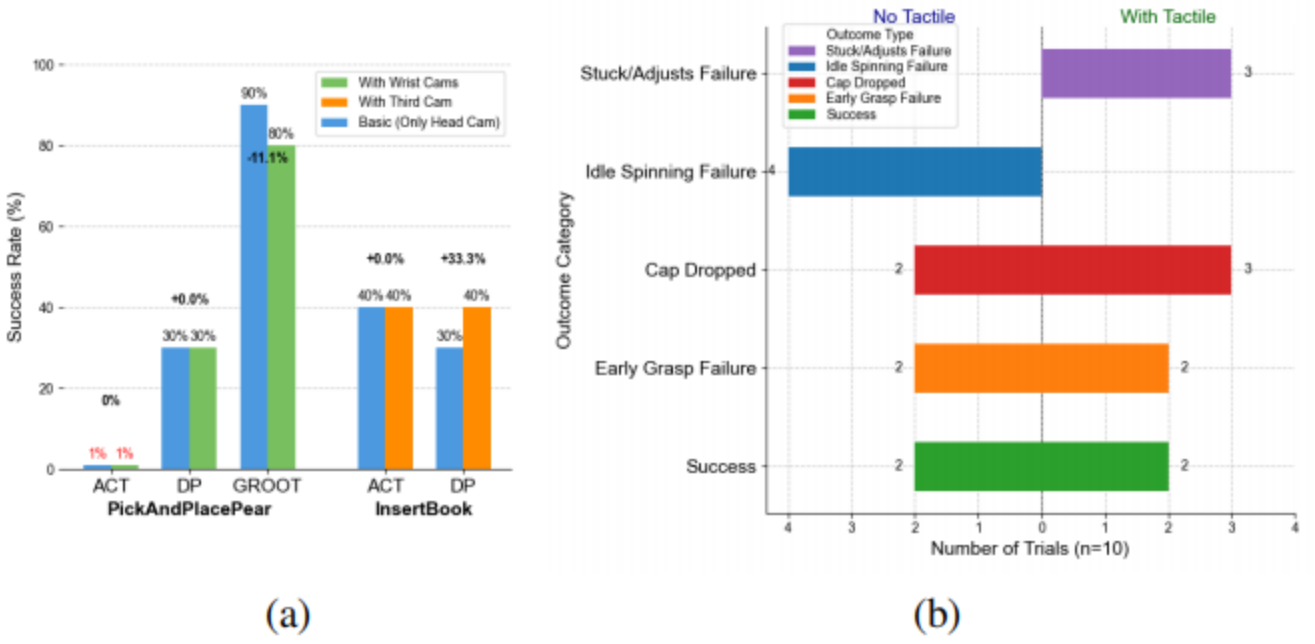

(a) (b)

Fig. 7. Ablation Study on Multi-view Configurations and Tactile Sensing. (a) We compares the success rates of different manipulation models on `PickAndPlacePear` and `InsertBook` under various observation configurations. (b) The distribution of various trial outcome for the `Unscrew Bottle` task contrast performance "No Tactile" feedback (left side) against "With Tactile" feedback (right side) over 10 trials.

We then ablate the tactile sensing by comparing DP's performance with and without it. We incorporate tactile information into observations as part of the low-dimensional state and test performance on `UnscrewBottle`, which is tactile-rich. We categorized the task outcomes into the following failure modes: *Stuck/Adjusts Failure*, where the right hand becomes stuck or fails to properly adjust its grip on the cap; *Idle Spinning Failure*, where the right hand spins without successfully engaging the cap; *Cap Dropped*, indicating that the cap is dropped after being unscrewed; and *Early Grasp Failure*, where the left hand fails to securely grasp the bottle body at the initial stage. As shown in Figure 7b, although success rates don't improve, the failure patterns shift from *Idle Spinning Failure* without tactile sensing to *Stuck/Adjusts Failure* when tactile feedback is incorporated. This indicates that with the addition of tactile feedback, the hand can better grasp the cap. This outcome underscores the unique value of tactile sensing: it reveals latent state variables that are visually ambiguous or entirely hidden, especially when the visual scene remains static but the physical contact state changes dramatically. This also demonstrates the significant value of multi-modal manipulation data.

## VI. Discussion, Limitations, and Future Work

We introduce a large-scale, multi-view, multi-modal humanoid upper-body manipulation dataset named RoboTacDex, encompassing diverse human-like manipulation tasks. Leveraging this dataset, we conduct systematic policy training and perform a detailed analysis of these models' strengths, weaknesses, and the types of tasks where they exhibit superior performance, as well as an ablation study on multi-view and multi-modality inputs. Additionally, we present a high-performance hardware integration and data collection system that resolves synchronization challenges across heterogeneous sensors. We believe RoboTacDex will serve as a valuable resource for advancing robust and generalizable humanoid manipulation systems.

One limitation of this study is that tactile information is incorporated in a relatively straightforward manner, without dedicated architectural components for multimodal fusion. Future work will explore more advanced model designs to better exploit tactile and multi-view observations. Furthermore, while the dataset spans diverse manipulation tasks, increasing its scale and environmental diversity remains an important direction for continued development.